\documentclass[letterpaper]{article} 
\usepackage{amsmath}
\usepackage{amsfonts}
\usepackage{aaai25}  
\usepackage{times}  
\usepackage{helvet}  
\usepackage{courier}  
\usepackage[hyphens]{url}  
\usepackage{graphicx} 
\usepackage{natbib}  
\usepackage{caption} 
\usepackage{pifont}

\usepackage{textcomp}
\usepackage{colortbl}
 \usepackage{multirow} 
\definecolor{abstractbg}{rgb}{1,0.969,0.914}
\usepackage{algorithm}
\usepackage{algorithmic}

\usepackage{newfloat}
\usepackage{listings}
\DeclareCaptionStyle{ruled}{labelfont=normalfont,labelsep=colon,strut=off} 
\floatstyle{ruled}
\newfloat{listing}{tb}{lst}{}
\floatname{listing}{Listing}
\title{FedPIA - Permuting and Integrating Adapters leveraging Wasserstein Barycenters for Finetuning Foundation Models in Multi-Modal Federated Learning}
\author{Pramit Saha\textsuperscript{\rm 1}, Divyanshu Mishra\textsuperscript{\rm 1}, Felix Wagner\textsuperscript{\rm 1}, Konstantinos Kamnitsas\textsuperscript{\rm 1,\rm 2,\rm 3}, J. Alison Noble\textsuperscript{\rm 1}
}
\affiliations{\textsuperscript{\rm 1} Department of Engineering Science, University of Oxford, \textsuperscript{\rm 2} Department of Computing, Imperial College London, \\
\textsuperscript{\rm 3} School of Computer Science, University of Birmingham \\
\{pramit.saha, divyanshu.mishra, felix.wagner, konstantinos.kamnitsas, alison.noble\}@eng.ox.ac.uk
}

\usepackage{bibentry}

\begin{document}

\maketitle

\begin{abstract}
Large Vision-Language Models (VLMs), possessing millions or billions of parameters, typically require large text and image datasets for effective fine-tuning. However, collecting data from various sites, especially in healthcare, is challenging due to strict privacy regulations. An alternative is to fine-tune these foundation models on end-user devices, such as in medical clinics and hospitals, without sending data to a server. These local clients typically have limited computing power and small datasets, which are not enough for fully fine-tuning large VLMs on their own. A naive solution to these scenarios is to leverage parameter-efficient fine-tuning (PEFT) strategies such as adapters and apply federated learning (FL) algorithms to combine the learned adapter weights, thereby respecting the resource limitations and data privacy of the clients. However, this approach does not fully leverage the knowledge from multiple adapters trained on diverse data distributions and for diverse tasks. The adapters are adversely impacted by data heterogeneity and task heterogeneity across clients resulting in sub-optimal convergence. To this end, we propose a novel framework called \textbf{FedPIA} that improves upon the naive combinations of FL and PEFT by introducing \textbf{P}ermutation and \textbf{I}ntegration of the local \textbf{A}dapters in the server and global \textbf{A}dapters in the clients exploiting Wasserstein barycenters for improved blending of client-specific and client-agnostic knowledge. This layerwise permutation helps to bridge the gap in the parameter space of local and global adapters before integration. We conduct over 2000 client-level experiments utilizing 48 medical image datasets across five different medical vision-language FL task settings encompassing visual question answering as well as image and report-based multi-label disease detection. Our experiments involving diverse client settings, ten different modalities, and two VLM backbones demonstrate that FedPIA consistently outperforms the state-of-the-art PEFT-FL baselines. Code/FL set up: \color{magenta}\url{https://github.com/PramitSaha/Fed-PEFT}\color{black}.

\end{abstract}

\vspace{-5mm}
\section{Introduction}
\label{sec:introduction}
Large Vision-Language Models (VLMs) have recently achieved significant progress in multi-modal learning \cite{NEURIPS2021_50525975,kim2021vilt}. These models excel at integrating and processing information from both image and text modalities, achieving impressive results when fine-tuned for real-world multi-modal tasks, such as Visual Question Answering (VQA) \cite{antol2015vqa} and Visual Commonsense Reasoning (VCR) \cite{zellers2019recognitioncognitionvisualcommonsense}.  Central to their capabilities are vast numbers of parameters, often in the millions or billions, which encapsulate the learned representations necessary for multi-modal comprehension. The evolution of large VLMs has highlighted the importance of fine-tuning, essential for adapting these models to specific tasks with high accuracy. Fine-tuning involves adjusting the model's parameters based on task-specific data, enhancing performance in targeted applications. To enhance the generalization ability of foundation models, extensive fine-tuning with large amounts of diverse data from various sources is typically required. However, aggregating all training data for centralized fine-tuning poses significant challenges. For example, collecting data from clinical centers across multiple countries is often infeasible due to privacy regulations.


\begin{figure*}
\vspace{-5mm}
    \centering
\includegraphics[width=1.8\columnwidth]{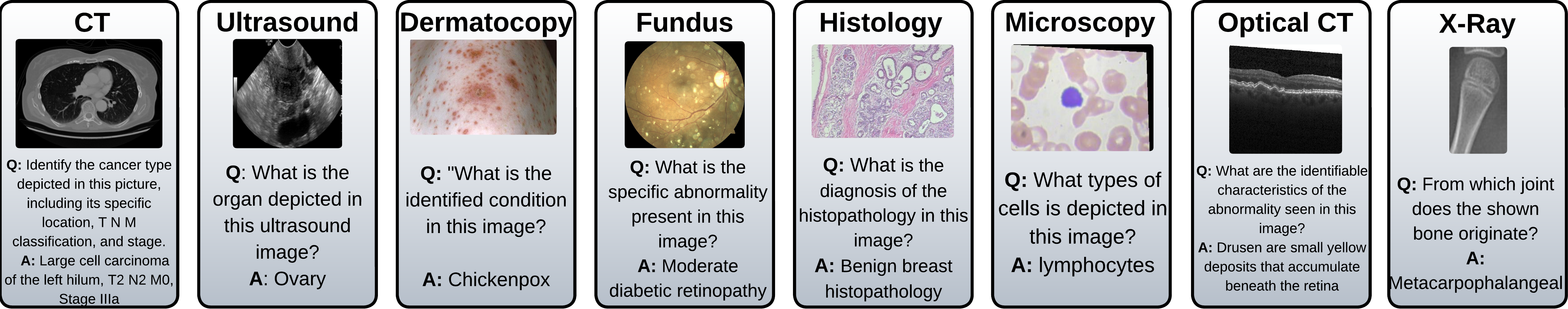}
    \caption{Sample VQA triplets of 8 modality-specific medical clients \textbf{(\color{cyan}{Task 2}\color{black} )}}
\label{fig1}
\vspace{-4mm}
\end{figure*}
The need to protect data privacy has led to the exploration of alternative approaches such as Federated Learning (FL) \cite{mcmahan2017communication,li2020federated,acar2021federated,karimireddy2020scaffold,saha2024f3ocusfederatedfinetuning,wagner2024feasibilityfederatedlearningclient,bdcc8090099,wagner2023post,saha2023rethinking}. FL involves training models on local devices, like those in medical clinics and hospitals, without transferring sensitive data to a central server. This decentralized approach mitigates the risk of data breaches and ensures compliance with privacy regulations. However, fine-tuning models in local devices faces challenges if there are limited computational resources and small, localized datasets. These limitations hinder the independent fine-tuning of large VLMs, which require extensive parameters and diverse datasets to capture the complexities of real-world language and visual data. Addressing these challenges requires solutions that balance data privacy with the computational and dataset limitations of local devices. 

Parameter-Efficient Fine-Tuning (PEFT) has recently gained attention in both Vision and Natural Language Processing (NLP) fields  \cite{hu2022lora,rebuffi2018efficient,lian2022scaling,ben-zaken-etal-2022-bitfit,frankle2021trainingbatchnormbatchnormexpressive,touvron2022three,lester2021power,VPT_jia,li2021prefix}. PEFT involves freezing the original backbone of the model and fine-tuning a small subset or a newly introduced set of parameters. FL, combined with PEFT, emerges as a promising paradigm for collaborative model training across decentralized clients while respecting data privacy and minimizing communication overhead.

Related works have primarily explored combinations of centralized PEFT algorithms and FedAvg. For example, some approaches focus on training and communicating adapters \cite{houlsby2019parameter} or a small number of trainable input tokens \cite{guo2023pfedprompt,guo2023promptfl}. These investigations are mostly limited to single modality scenarios, addressing only visual or textual tasks. Besides, none of these studies tackle the issue of data heterogeneity or task heterogeneity that lead to model drifts during local client updates and result in an unstable and sub-optimal convergence of the server model \cite{li2020federated}. Besides, such naive combination of FL with PEFT does not fully leverage the knowledge embedded in the multiple models trained on heterogeneous data distributions and diverse tasks. The recent work FedDAT \cite{chen2024feddat} leverages a Dual-Adapter Teacher (DAT) module, consisting of two parallel adapters: a local adapter and a frozen global adapter. However, the local adapters are trained independently in different clients per round and involve individual complex information streams, thereby making them distant in the parameter space.
This implies that mere addition of the diverse global and local adapters without proper alignment leads to sub-optimal performance and catastrophic forgetting, particularly in data- and task-heterogeneous FL. Besides, it requires the use of a separate global adapter teacher along with alternate training of the DAT module via mutual knowledge distillation that adds to its training complexity.

To address this issue, we present a novel framework called FedPIA (\textbf{Fed}erated Learning via \textbf{P}ermuting and \textbf{I}ntegrating \textbf{A}dapters) to improve information sharing between the client adapters in the server as well as between the local and global adapters in the clients. This is achieved by: (a) bringing the client adapters closer to each other in the server and (b) bringing the global adapters closer to the client-specific adapters in the clients in the parameter space. Concretely, in the server, in each layer, we permute the diverse, client adapter neurons to match the initialized global adapter neurons (obtained via FedAvg \cite{mcmahan2017communication}), before combining them, utilizing the theory of Wasserstein Barycenters. Furthermore, in order to better integrate client-specific and client-agnostic knowledge in the clients, we permute the weights of the global adapter in each client and bring it closer to client-specific adapter in the weight space before combining them. This two-fold approach bridges the gap between diverse adapters which are originally optimized on different distributions, using different input features from different modalities and leads to a stable convergence as seen in Fig 4. In order to showcase the effectiveness of FedPIA, we carry out over 2000 client-level experiments under five Vision-language FL task settings using 48 medical image datasets that involves data heterogeneity, modality heterogeneity, and task heterogeneity. The results demonstrate that FedPIA shows consistent and robust performance irrespective of heterogeneity conditions, outperforming the baselines for all task scenarios.

\begin{figure*}
\vspace{-5mm}
    \centering
\includegraphics[width=1.8\columnwidth]{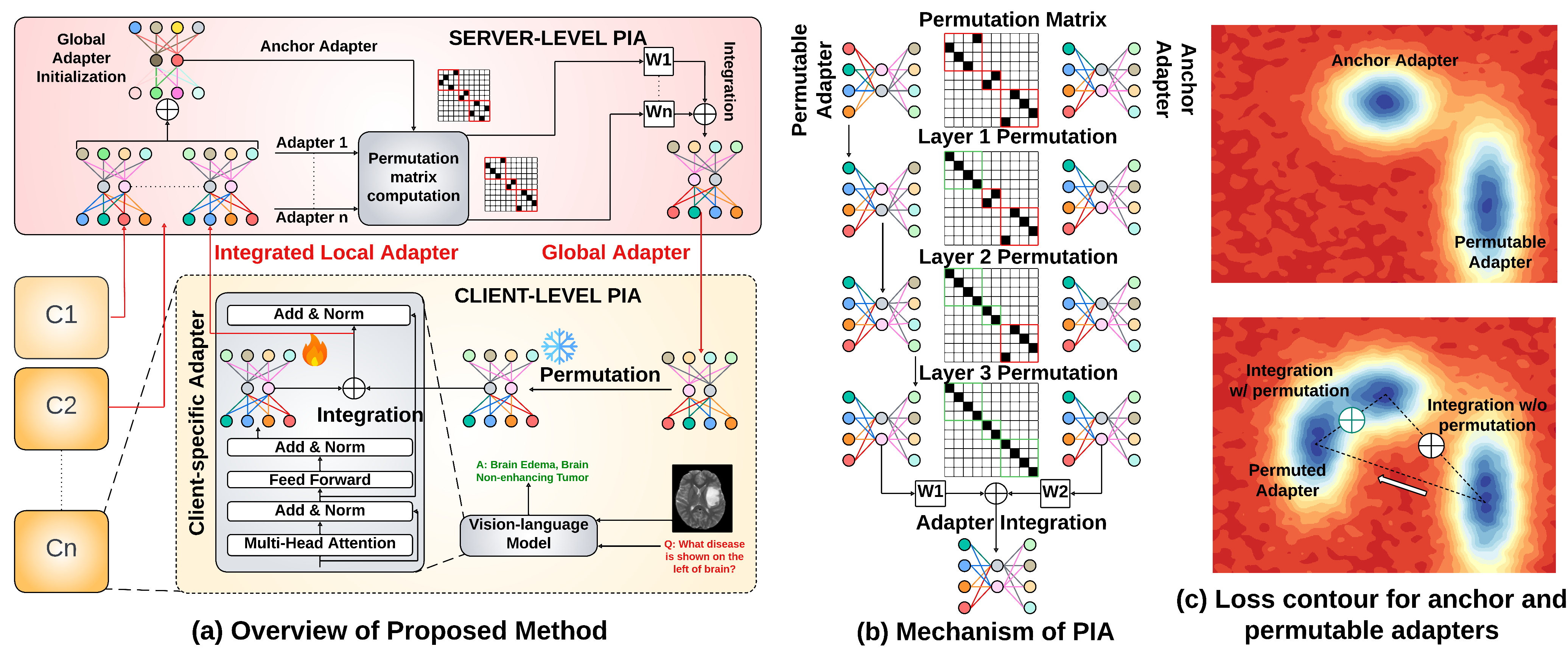}
    \caption{Illustration of FedPIA: (a) shows the permutation of integrated client adapter weights to match the initialized global adapter (anchor) in server as well as the permutation of global adapter weights to match the client-specific adapter (anchor) in clients, (b) shows the layerwise computation of permutation matrix between two adapters followed by their integration, (c) shows the motivation behind PIA in the loss landscape. The permutation matrix computed in (b) is used to project the permutable adapter into the same loss basin as the anchor adapter which leads to improved convergence of the integrated adapter.}
\label{fig1}
\vspace{-4mm}
\end{figure*}

\section{Background and Related Work}

\subsubsection{Federated Learning (FL)}
FL enables various clients to collaboratively train models in a decentralized manner without sharing local data. The classical FL framework, FedAvg \cite{mcmahan2017communication}, offers a practical method for model aggregation. 
However, its performance is adversely impacted by client non-IID data distributions. Consequently, several modifications have emerged to address data heterogeneity \cite{li2020federated,karimireddy2020scaffold,acar2021federated}.  
FedProx \cite{li2020federated} adds a proximal term to the client loss function thereby enforcing constraints on local updates. Another work called Scaffold \cite{karimireddy2020scaffold} employs control variates to enhance local updates, while FedDyn \cite{acar2021federated} dynamically regulates the client loss function to align the local and the global objectives. Moon \cite{li2021model} regularizes local training via contrastive learning. All these works assume unimodal data in all clients.

\subsubsection{Parameter-efficient Fine-tuning (PEFT)}
PEFT techniques can be categorized into three families: adaptive methods, selective methods, and prompt tuning. 
Adaptive methods are additive PEFT techniques that integrate adapters or small neural network blocks into the Transformer layers \cite{hu2022lora,rebuffi2018efficient,li2022cross,lian2022scaling}. 
Selective PEFT fine-tunes a subset of the existing parameters to enhance model performance on downstream tasks \cite{ben-zaken-etal-2022-bitfit,frankle2021trainingbatchnormbatchnormexpressive,touvron2022three}. 
Prompt tuning methods modify the original input, whether an embedding or the actual instance, with some prompts consisting of additional trainable parameters or perturbations \cite{lester2021power,VPT_jia,li2021prefix}.


\subsubsection{PEFT and Federated Learning}

The application of PEFT in multimodal FL remains relatively unexplored. Previous research has primarily adapted PEFT for FL in a straightforward manner, particularly focusing on uni-modal tasks, \textit{i.e.}, vision or NLP. 
\cite{chen2022fedtune} and \cite{sun2022exploring} evaluate existing PEFT baselines combined with FL in vision tasks. \cite{guo2023pfedprompt}, \cite{guo2023promptfl}, \cite{li2023visual}, and \cite{lu2023fedclip} fine-tune CLIP by communicating a small amount of learnable personalized prompts. \cite{su2022cross} addresses the issue of heterogeneous client images by adding adapters. \cite{yang2024exploring} explores the possibility of fine-tuning diffusion models via FL.
\cite{yu2023federated} optimize adapters for few-shot fine-tuning of LLMs. \cite{zhang2024towards} builds distributed instruction tuning datasets and fine-tunes a LLM via Low-Rank Adaptation \cite{hu2022lora}. \cite{zhuang2023foundation} analyzes the challenges of fine-tuning LLMs in FL. 

\cite{yu2023multimodal} is the first work to consider multi-modal client datasets. However, it processes visual and language data using separate networks, without utilizing a unified VLM. A recent work \cite{nguyen2024flora} proposes FLORA for fine-tuning VLMs using LoRA adapters in FL. \cite{zeng2024open} introduces a multimodal prototyping mechanism for fine-tuning VLMs. FedDAT \cite{chen2024feddat} considers data heterogeneity in multimodal FL by utilizing a Dual-Adapter Teacher (DAT) and employing Mutual Knowledge Distillation (MKD) between the local and global adapters. It is the only work on federated PEFT of VLMs for VQA. However, as discussed earlier, FedDAT does not fully utilize the knowledge embedded in multiple local adapters trained on heterogeneous data distributions and diverse tasks. It needs a separate global adapter for MKD thereby doubling the total number of trainable parameters. We tackle the heterogeneity issue without adding any training overhead.

\section{Methodology: FedPIA}
\subsubsection{Problem definition:}
We tackle a heterogeneous FL problem involving \( K \) clients. Each client \( k \) possesses a private multimodal dataset \( D_k \), which includes both visual (\( v_k \)) and textual (\( t_k \)) data. 
Specifically, each local dataset \( D_k \) can be decomposed into \( N_k \) image-text-output triplets \(\{(v_{k_i}, t_{k_i}, a_{k_i}) | i \in \{1, \ldots, N_k\}\}\). We assume that the marginal distribution of \( v_{k_i} \), \( t_{k_i}\), and \(a_{k_i} \) varies across clients, indicating data heterogeneity in the visual, textual and task domains. We define the answer or label pool \( A_k = \{a_{k_1}, \ldots, a_{k_{C_k}}\} \) with \( C_k \) ground-truth answers or labels for client \( k \), and frame our task as a \( C_k \)-way classification problem. Note that the answer pool and the total number of classes differ from client to client, thereby inducing heterogeneity in the FL model. 
Let \( f \) be a foundation model parameterized by \( \theta \). Starting from the pre-trained weights $\theta_{0}$, the goal is to optimize client-specific losses $L_k$ by gradient descent. Due to client-specific data and resource constraints, full fine-tuning is not feasible in FL. 
Our goal is to collaboratively fine-tune the foundation model \( f_\theta \) in a parameter-efficient manner within a predefined communication budget. For this, following additive PEFT, we introduce new parameters \( \phi \) for fine-tuning while keeping the original model frozen, resulting in the full parameter set \( \theta' = \{\theta, \phi\} \). 
\subsubsection{Overall idea:} The client adapters communicated to the server are distant in the weight space due to heterogeneity in client task space and data distribution, as indicated by the convergence analysis in Fig. 4 (Detailed analysis in \textbf{Suppl. \S C}). Owing to the permutation invariance property, these adapters lack one-to-one correspondence, which is crucial for effective information fusion. Therefore, we adopt the theory of Wasserstein Barycenters \cite{singh2020model,akash2022wasserstein} to synchronise and combine multiple client adapters in layerwise fashion and using weight space as their underlying distribution in the server. The Wasserstein Barycenter relates to the concept of averaging in the Wasserstein space by minimizing the Earth Mover's distance between the barycenter and given distributions. This helps in bringing the adapters closer in the parameter space, prior to aggregation, as seen in Fig. 2(c). The aggregated global adapter is communicated back to each client. However, it possesses client-agnostic knowledge and is again distant from the local adapters in the weight space. Using the global adapter in this form leads to slower, unstable convergence, as seen in Fig. 4(a). Therefore, we permute the global model in each client to match the local adapter before integration using similar technique. This permuted global adapter is consequently frozen and combined with the client-specific adapter, thereby integrating client-specific and shared knowledge as seen in Fig. 2(a). At the end of each round, this integrated adapter (also called client adapter) is uploaded to the server (see \textbf{Algorithm 1} in \textbf{Suppl. \S A}). 
\subsubsection{Server-level PIA:}

We introduce a two-step procedure in the server: First, we initialize the global adapter using standard FedAvg \cite{mcmahan2017communication} of the client adapters. Next, we permute each client adapter to match the initialized global adapter by computing the permutation matrix as observed in Fig. 2. For this, we define probability measure over neurons in the $l^{th}$ layer for the $k^{th}$ client adapter as \(\mu_k^{(\ell)} = (\alpha_k^{(\ell)}, X_k[\ell])\) and that for the estimated global adapter as \(\nu^{(\ell)} = (\beta^{(\ell)}, X_\mathcal{G}[\ell])\), where \(X_k\) and \(X_\mathcal{G}\) are the respective measure supports. The weight \(\alpha = (\alpha_1, \ldots, \alpha_n)\) lies in the probability simplex \(\Sigma_n := \{a \in \mathbb{R}^n_+ \mid \sum_{i=1}^n a_i = 1\}\) (and similarly for \(\beta\)).
We consider that the support of each adapter neuron in the server is given by the weights of the incoming edges, which are stacked in a vector. Accordingly, an adapter neuron can be represented by the corresponding row in the weight matrix. Therefore, the support of their measures is given by \(X_k[\ell] = W^{(\ell, \ell-1)}_k\) and \(X_{\mathcal{G}}[\ell] = W^{(\ell, \ell-1)}_{\mathcal{G}}\). 

Let \(C_k^{ij, (l)}\) denote the ground cost of permuting the $i^{th}$ adapter neuron of $l^{th}$ layer in the $k^{th}$ client to the $j^{th}$ adapter neuron of same layer in the server. It is equivalent to moving the measure supports from \(X_{k,(i)}[l]\) to \(X_{\mathcal{G},(j)}[l]\).
We compute this cost as the Euclidean distance between the weight vector of the local and initialized global adapter, \textit{i.e.},  $C_k^{(l)} = \| \mathbf{X}_k^{i}[\ell] - X_{\mathcal{G}}^{j}[\ell] \|_2, \quad \forall i \in [n_k^{(\ell)}], j \in [m^{(\ell)}]$ where \( n_k^{\ell} \) and \( m^{\ell} \) are the number of neurons in the $\ell^{th}$ layer of the client adapter and the global adapter respectively.
We initialize the probability mass values of each layer from a uniform distribution. So, \( \alpha_k^{(\ell)} = \frac{1}{n^{(\ell})}, \beta^{(\ell)} = \frac{1}{m^{(\ell)}} \). 

For aligning the incoming weights $W_k^{(\ell, \ell-1)}$ for the $l^{th}$ layer in $k^{th}$ client adapter, we first normalize the previous layer permutation matrix $P_k^{(l-1)}$ with the inverse of corresponding column marginals of the server adapter $\beta ^ {(l-1)}$ as $P_k^{(\ell-1)} \text{diag} (1/\beta^{(\ell-1)})$ and post-multiply with the current layer weights of the client adapter $W^{(\ell, \ell-1)}_k$. 
Next, based on the cost metric \(C_k^{(l)}\), we compute the permutation matrix $P_k^\ell$ between measures $\mu_k^\ell,\nu^\ell$ for the current layer ($\ell$) by minimizing the Wasserstein distance $\mathcal{W}^{(\ell)}\left(\mu_k^{(\ell)}, \nu^{(\ell)}, C_k^{(l)}\right)$. This permutation matrix is used to align the the client adapter and global adapter weights as:
\begin{equation}
\widetilde{\mathbf{W}}_k^{(\ell, \ell-1)} = \text{diag}\left(1/\beta^{(\ell)}\right) P^{(\ell)\top}_k W^{(\ell, \ell-1)}_k P_k^{(\ell-1)} \text{diag} (1/\beta^{(\ell-1)})
\end{equation}
The aligned $K$ adapters are then integrated dynamically to form the global adapter as: $\widetilde{\mathbf{W}}_\mathcal{G}^{(\ell, \ell-1)} = \frac{1}{K} \sum_{k=1}^{K} \widetilde{\mathbf{W}}_k^{(\ell, \ell-1)} \exp \left( -\gamma\lVert \widetilde{\mathbf{W}}_k - \mathbf{W}_\mathcal{G} \rVert_2 \right)$ where $\gamma$ is a hyperparameter. Note that our method is orthogonal to different FL aggregation schemes and can be used in conjunction with those. For simplicity, we use FedAvg as the chosen aggregation scheme in this work without loss of generality.

\subsubsection{Client-level PIA:} In the clients, we combine the global adapter and client-specific adapter (with parameters from the last communication round) for integrating client-specific and shared information streams. For this, we first align the global adapter $\widetilde{\mathbf{W}}_\mathcal{G}$ to the client-specific adapter $\mathbf{W}_k$ using the Wasserstein distance following the aforementioned procedure. The only difference in the permutation computation here is that we use adapter activations rather than weights for computing the cost metric \(C_k^{(\ell)}\). For this, we compute the mean neuron activation ($\psi$) for all the neurons of local and global adapters over a randomly selected batch of $m$ samples $B = \{x\}_{i=1}^{m}$ and use it as the support of measures. Therefore, the support is now denoted as \(X_k [\ell] = \psi (M_k^{\ell} (B))\) and \(X_{\mathcal{G}} [\ell] = \psi (M^{\ell}_{\mathcal{G}} (B))\) for the local and global adapter respectively, where $M$ denotes the adapter model. In other words, neurons across local and global adapters in each client would be considered similar if they yield similar activations for a given instance. Leveraging the activations instead of weight matrix particularly helps the global adapter to adapt to the client-specific data distribution better (as indicated in Table 5) since the activations (unlike weights) are directly dependent on the input data.

\begin{table*}[t]
    \centering
    \caption{Performance comparison of FedPIA with other methods on \color{magenta}{Task 1} \color{black}(in terms of accuracy)}
    \vspace{-3mm}
    \scalebox{0.6}{
    \begin{tabular}{l|lll|lll|lll|lll|lll|lll}
        \hline
        \textbf{Fine-tuning} & \multicolumn{3}{c|}{\cellcolor[rgb]{ .886,  .937,  .855}\textbf{Slake}} & \multicolumn{3}{c|}{\cellcolor[rgb]{ .886,  .937,  .855}\textbf{VQA-Med 2019}} & \multicolumn{3}{c|}{\cellcolor[rgb]{ .886,  .937,  .855}\textbf{VQA-Med 2020}} & \multicolumn{3}{c|}{\cellcolor[rgb]{ .886,  .937,  .855}\textbf{VQA-Med 2021}} & \multicolumn{3}{c|}{\cellcolor[rgb]{ .886,  .937,  .855}\textbf{VQA-RAD}} & \multicolumn{3}{c}{\cellcolor[rgb]{ .706,  .776,  .906}\textbf{Mean Score}} \\
        
        \hline
        & \textbf{Open} & \textbf{Closed} & \textbf{Overall} & \textbf{Open} & \textbf{Closed} & \textbf{Overall} & \textbf{Open} & \textbf{Closed} & \textbf{Overall} & \textbf{Open} & \textbf{Closed} & \textbf{Overall} & \textbf{Open} & \textbf{Closed} & \textbf{Overall} & \textbf{Open} & \textbf{Closed} & \textbf{Overall}\\
        \hline \hline
        \rowcolor[rgb]{ .988,  .894,  .839}\textbf{} & \multicolumn{18}{c}{\textbf{Backbone architecture: ViLT}}\\
        \hline
        Full fine-tuning & 74.73 & 74.48 & 74.66 & 60.55 & 59.38 & 60.43 & 0.70 & 52.94 & 15.54 & 21.00 & ~~~~- & 21.00 & 42.47 & 64.26 & 55.90 & 39.89 & 62.77 & 45.51\\ \hline
        Classifier only (LB) & 66.82 & 60.82 & 65.60 & 54.13 & 50.00 & 53.72 & 0.00 & 37.25 & 10.88 & 18.50 & ~~~~- & 13.00 & 34.95 & 57.41 & 47.88 & 37.06 & 52.10 & 38.47\\
        AdapterFusion & 72.87 & 68.75 & 72.10 & 57.57 & 54.69 & 57.28 & 0.70 & {52.94} & {15.54} & 19.50 & ~~~~- & 19.50 & 33.33 & 58.55 & 49.00 & 36.79 & 58.73 & 42.68\\
        Houlsby Adapter & {73.80} & 71.64 & {73.61} & 57.80 & 57.81 & 57.80 & 0.00 & {52.94} & 14.51 & 20.50 & ~~~~- & 20.50 & 30.65 & 59.32 & 48.78 & 36.55 & {60.43} & 43.04\\
        Parallel Adapter & 72.71 & 64.18 & 70.50 & {58.26} & 57.81 & {58.21} & 0.00 & 49.02 & 13.99 & 24.00 & ~~~~- & 24.00 & 32.80 & 54.75 & 46.55 & 37.55 & 56.44 & 42.65\\
        Compacter & 71.68 & 65.39 & 70.31 & {58.26} & 57.81 & {58.21} & 0.00 & 50.98 & 14.51 & 22.00 & ~~~~- & 22.00 & 39.95 & 57.03 & 48.55 & 38.38 & 57.80 & 42.72\\
        LayerNorm & 72.40 & 64.90 & 70.59 & 56.65 & 48.44 & 55.83 & 0.70 & 33.33 & 9.84 & 17.00 & ~~~~- & 17.00 & 32.26 & 57.03 & 47.44 & 35.80 & 50.93 & 40.14\\
        LoRA & 60.93 & 56.97 & 59.94 & 53.90 & 56.25 & 55.80 & 0.00 & 15.69 & 4.66 & 15.00 & ~~~~- & 15.00 & 23.12 & 56.65 & 43.88 & 30.59 & 46.39 & 35.86\\
        Bias & 73.33 & 66.83 & 71.72 & 57.80 & 46.88 & 56.71 & 0.70 & 37.25 & 10.88 & 18.50 & ~~~~- & 18.50 & 34.95 & 57.41 & 48.55 & 37.06 & 52.10 & 41.27\\
        PromptFL (k=5) & 63.41 & 57.21 & 62.02 & 55.27 & 51.56 & 54.90 & 0.00 & 15.64 & 4.15 & 16.00 & ~~~~- & 16.00 & 27.96 & 58.17 & 46.55 & 32.53 & 45.65 & 36.72\\
        PromptFL (k=10) & 71.32 & 67.55 & 70.31 & 56.88 & 48.88 & 56.08 & 0.00 & 11.76 & 3.63 & 19.00 & ~~~~- & 19.00 & 25.27 & 58.18 & 45.21 & 34.49 & 46.59 & 38.85\\
        PromptFL (k=20) & 65.74 & 68.02 & 67.48 & 56.42 & 51.56 & 55.93 & 0.70 & 11.77 & 4.15 & 20.00 & ~~~~- & 20.00 & 27.96 & 58.56 & 46.77 & 34.16 & 47.48 & 38.87\\
        PromptFL (k=50) & 71.01 & 66.83 & 70.22 & 57.34 & 57.81 & 57.39 & 0.70 & 7.84 & 3.11 & 19.00 & ~~~~- & 19.00 & 25.27 & 55.54 & 43.88 & 34.66 & 47.01 & 38.72\\
        FedDAT (AdapterFusion) & 72.09 & 69.71 & 71.91 & 55.96 & 57.81 & 56.15 & 0.00 & 31.37 & 8.81 & 23.50 & ~~~~- & 23.50 & {42.47} & {62.36} & {55.01} & 38.80 & 55.31 & {43.08}\\
        FedDAT (Houlsby) & 73.02 & 73.07 & 73.03 & 53.21 & 51.56 & 53.05 & 0.00 & 45.09 & 12.44 & 23.00 & ~~~~- & 23.00 & 38.71 & 56.65 & 50.33 & 37.59 & 56.59 & 42.54\\
        FedDAT (Parallel) & 72.87 & 70.91 & 72.14 & 56.65 & 50.00 & 55.99 & 0.00 & 29.41 & 7.77 & {25.00} & ~~~~- & {25.00} & 39.79 & 58.56 & 51.89 & {38.86} & 52.22 & 42.56\\
        FedDAT (Compacter) & 73.02 & {74.04} & 73.28 & 56.57 & {59.38} & 56.85 & 0.00 & 41.18 & 11.92 & 24.50 & ~~~~- & 24.50 & 38.17 & 52.09 & 47.66 & 38.45 & 56.67 & 42.86\\
    
        FedPIA (Houlsby) \textbf{(ours)} & \textbf{74.45} & \textbf{74.20} & \textbf{74.38} & 60.29 & 60.16 & 60.26 & 0.70 & \textbf{53.20} & \textbf{15.86} & \textbf{27.50} & ~~~~- & \textbf{27.50} & 42.67 & 64.05 & 55.37 & \textbf{41.12} & 62.90 & \textbf{46.66}\\
        FedPIA (Parallel) \textbf{(ours)} & 74.04 & 74.16 & 74.10 & 59.90 & 59.82 & 59.87 & 0.70 & 52.48 & 15.36 & 27.00 & ~~~~- & 27.00 & 42.01 & 64.02 & 55.12 & 40.73 & 62.62 & 46.29\\
        FedPIA (Compacter) \textbf{(ours)} & 74.41 & 74.18 & 74.35 & \textbf{60.35} & \textbf{60.86} & \textbf{60.42} & 0.70 & 52.88 & 15.49 & 27.00 &  ~~~~- & 27.00 & \textbf{43.16} & \textbf{64.33} & \textbf{56.04} & \textbf{41.12} & \textbf{63.06} & \textbf{46.66}\\
        \hline
       \rowcolor[rgb]{ .988,  .894,  .839} \textbf{} & \multicolumn{18}{c}{\textbf{Backbone architecture: ALBEF}}\\
        \hline
       Full fine-tuning & 77.89 & 77.28 & 77.45 & 67.65 & 63.20 & 67.25 & 0.70 & 50.08 & 15.06 & 22.00 & ~~~~- & 22.00 & 39.35 & 62.22 & 52.86 & 41.52 & 63.20 & 46.92 \\ \hline
        Classifier only (LB) & 70.23 & 65.81 & 69.24 & 61.03 & 58.45 & 60.73 & 0.00 & 34.27 & 9.85 & 19.00 & ~~~~- & 19.00 & 30.94 & 56.76 & 44.59 & 36.24 & 53.82 & 40.68 \\
        AdapterFusion & 72.96 & 70.37 & 72.46 & 64.20 & 60.77 & 63.29 & 0.00 & 40.26 & 11.39 & 19.50 & ~~~~- & 19.50 & 31.43 & 58.78 & 45.76 & 37.62 & 57.55 & 42.48 \\
        Houlsby Adapter & 73.00 & 72.44 & 72.82 & 64.85 & 61.25 & 64.45 & 0.00 & 39.70 & 11.56 & 21.00 & ~~~~- & 21.00 & 32.88 & 57.02 & 46.03 & 38.35 & 57.60 & 43.17 \\
        Parallel Adapter & 72.10 & 67.54 & 70.67 & 65.39 & 60.67 & 64.78 & 0.00 & 41.55 & 12.04 & 24.50 & ~~~~- & 24.50 & 33.89 & 53.20 & 45.66 & 39.18 & 55.74 & 43.53 \\
        Compacter & 72.19 & 68.30 & 71.18 & 64.26 & 58.20 & 63.85 & 0.00 & 42.88 & 12.46 & 21.00 & ~~~~- & 21.00 & 35.63 & 56.95 & 47.98 & 38.62 & 56.58 & 43.29 \\
        LayerNorm & 70.27 & 66.03 & 69.53 & 63.43 & 51.12 & 62.22 & 0.70 & 35.92 & 10.59 & 22.00 & ~~~~- & 22.00 & 28.07 & 55.07 & 42.15 & 36.89 & 52.04 & 41.30 \\
        LoRA & 58.25 & 55.20 & 57.73 & 60.24 & 58.78 & 60.18 & 0.00 & 15.56 & 4.59 & 15.00 & ~~~~- & 15.00 & 22.30 & 52.84 & 37.94 & 31.16 & 45.60 & 35.09 \\
        Bias & 70.24 & 63.98 & 68.25 & 56.76 & 45.09 & 55.61 & 0.00 & 38.43 & 11.17 & 17.00 & ~~~~- & 17.00 & 33.37 & 54.46 & 45.69 & 35.47 & 50.49 & 39.54 \\
        PromptFL (k=5) & 67.21 & 60.24 & 65.63 & 58.52 & 50.06 & 57.56 & 0.70 & 13.95 & 4.78 & 15.00 & ~~~~- & 15.00 & 24.50 & 55.23 & 40.26 & 33.19 & 44.87 & 36.65 \\
        PromptFL (k=10) & 68.22 & 60.77 & 66.14 & 59.32 & 51.13 & 58.54 & 0.00 & 15.80 & 4.84 & 17.50 & ~~~~- & 17.50 & 26.34 & 53.32 & 40.53 & 34.28 & 45.26 & 37.51 \\
        PromptFL (k=20) & 67.93 & 63.34 & 66.61 & 61.23 & 57.74 & 60.89 & 0.00 & 12.22 & 3.41 & 20.00 & ~~~~- & 20.00 & 25.90 & 54.45 & 40.25 & 35.01 & 46.94 & 38.23 \\
        PromptFL (k=50) & 68.05 & 62.09 & 66.44 & 60.55 & 58.30 & 60.32 & 0.00 & 10.56 & 3.12 & 19.50 & ~~~~- & 19.50 & 25.82 & 53.56 & 39.88 & 34.78 & 46.13 & 37.85 \\
        FedDAT (AdapterFusion) & 72.48 & 70.76 & 72.12 & 63.36 & 59.85 & 62.95 & 0.70 & 42.66 & 12.68 & 20.50 & ~~~~- & 20.50 & 33.10 & 58.03 & 46.76 & 38.03 & 57.83 & 43.00 \\ 
        FedDAT (Houlsby) & 73.22 & 71.34 & 72.79 & 64.05 & 60.29 & 63.58 & 0.00 & 43.28 & 12.46 & 23.00 & ~~~~- & 23.00 & 34.28 & 57.97 & 47.82 & 38.91 & 58.22 & 43.93 \\ 
        FedDAT (Parallel) & 72.80 & 70.90 & 72.19 & 63.13 & 60.38 & 62.85 & 0.00 & 42.32 & 12.28 & 24.00 & ~~~~- & 24.00 & 33.69 & 59.75 & 47.79 & 38.72 & 58.34 & 43.82 \\ 
        FedDAT (Compacter) & 73.01 & 71.25 & 72.43 & 64.19 & 59.93 & 63.72 & 0.00 & 43.99 & 12.36 & 24.50 & ~~~~- & 24.50 & 32.90 & 58.44 & 46.73 & 38.92 & 58.40 & 43.95 \\ 
        FedPIA (Houlsby)  \textbf{(ours)}& 76.26 & 76.99 & 76.86 & \textbf{67.25} & 62.26 & 66.76 & 0.70 & 50.14 & 15.13 & 27.50 & ~~~~- & 27.50 & \textbf{42.39} & \textbf{62.80} & \textbf{54.97} & \textbf{42.82} & 63.05 & \textbf{48.24} \\ 
        FedPIA (Parallel) \textbf{(ours)} & \textbf{77.17} & 75.65 & 76.78 & 66.78 & \textbf{62.98} & \textbf{66.96} & 0.70 & 49.06 & 14.59 & \textbf{28.00} & ~~~~- & \textbf{28.00} & 41.13 & 62.33 & 53.93 & 42.76 & 62.51 & 48.05 \\ 
        FedPIA (Compacter) \textbf{(ours)}  & 76.80 & \textbf{77.12} & \textbf{76.89} & 67.13 & 62.84 & 66.75 & 0.70 & \textbf{51.25} & \textbf{15.45} & 27.50 & ~~~~- & 27.50 & 40.48 & 61.49 & 53.15 & 42.52 & \textbf{63.18} & 47.95 \\ 
        \hline
    \end{tabular}}
\end{table*}

\begin{table*}[t]
    \centering
    \caption{Comparison of FedPIA with other methods on \color{cyan}{Task 2} \color{black} with modality-specific clients (in terms of accuracy)}
    \vspace{-2mm}
    \scalebox{0.7}{
    \begin{tabular}{l|llllllll|l}
        \hline
        \textbf{Fine-tuning}  & \cellcolor[rgb]{ .886,  .937,  .855}\textbf{C1} & \cellcolor[rgb]{ .886,  .937,  .855}\textbf{C2} & \cellcolor[rgb]{ .886,  .937,  .855}\textbf{C3} & \cellcolor[rgb]{ .886,  .937,  .855}\textbf{C4} & \cellcolor[rgb]{ .886,  .937,  .855}\textbf{C5} & \cellcolor[rgb]{ .886,  .937,  .855}\textbf{C6} & \cellcolor[rgb]{ .886,  .937,  .855}\textbf{C7} & \cellcolor[rgb]{ .886,  .937,  .855}\textbf{C8} & \cellcolor[rgb]{ .706,  .776,  .906}\textbf{Overall} \\
        \ & {(CT)} & {(US)} & {(OCT)} & {(Fundus)} & {(Micro.)} & {(Hist.)} & {(Derma.)} & {(XRay)} &  \\
        \hline \hline
        Full fine-tuning & 94.95 & 87.63 & 93.55 & 81.84 & 92.65 & 93.00 & 76.12 & 92.20 & 88.99  \\\hline
        Classifier only (LB) & 80.30 & 75.17 & 73.76 & 69.99 & 86.32 & 87.96 & 68.11 & 84.96 &  78.32 \\
        AdapterFusion & 83.33 & 80.54 & 80.97 & 71.59 & 90.74 & 92.12 & 71.33 & 88.61 & 82.40  \\
        Houlsby Adapter & 82.58 & 80.49 & 83.44 & 68.86 & 89.85 & 92.34 & 74.18 & 87.92 & 82.46  \\
        Parallel Adapter & 79.29 & 79.76 & 83.33 & 68.02 & 90.29 & 91.90 & 71.03 & 89.37 & 81.63  \\
        Compacter & 83.33 & 80.08 & 80.86 & 73.47 & 93.24 & 91.25 & 71.63 & 86.89 & 82.59  \\
        LayerNorm & 81.82 & 77.94 & 78.17 & 72.72 & 89.26 & 91.25 & 69.24 & 87.23 & 80.95  \\
        LoRA & 58.84 & 61.53 & 62.04 & 67.73 & 73.82 & 89.72 & 59.73 & 80.54 & 69.24  \\
        Bias & 82.32 & 77.63 & 79.35 & 71.97 & 89.71 & 91.47 & 70.21 & 86.89 & 81.19  \\
        PromptFL & 79.80 & 71.35 & 78.39 & 71.59 & 89.26 & 93.00 & 70.88 & 85.44 & 79.96  \\

        FedDAT (AdapterFusion) & 82.07 & 79.99 & 79.35 & 74.41 & 91.32 & 89.72 & 71.63 & 88.44 & 82.37 \\
        FedDAT (Houlsby) & 79.80 & 79.22 & 76.67 & 75.07 & 90.29 & 90.59 & 72.90 & 89.86 & 81.55\\
        FedDAT (Parallel) & 82.83 & 81.26 & 80.54 & 74.60 & 89.41 & 92.12 & 73.35 & 88.79 & 82.61 \\
        FedDAT (Compacter) & 80.56 & 80.44 & 81.61 & 73.38 & 90.15 & 90.37 & 72.23 & 89.37 & 82.01 \\
    
        FedPIA (Houlsby) \textbf{(ours)} & 89.86 & 86.77 & \textbf{90.03} & 80.20 & \textbf{92.34} & \textbf{92.88} & 75.68 & \textbf{91.99} & \textbf{87.97}\\
        FedPIA (Parallel) \textbf{(ours)} & 88.94 & \textbf{87.00} & 88.59 & 79.97 & 90.80 & 90.93 & 74.32 & 90.02 & 86.82 \\
        FedPIA (Compacter) \textbf{(ours)} & \textbf{90.06} & 86.29 & 89.01 & \textbf{81.14} & 90.25 & 89.24 & \textbf{76.24} & 91.23 & 86.93 \\
        \hline
    \end{tabular}}
\end{table*}

\begin{table*}[htbp]
    \centering
    \caption{Performance comparison of FedPIA with other methods on Tasks \color{blue}3 \color{black} and \color{red}4 \color{black} using ViLT (in terms of F1 score)}
    \vspace{-2mm}
    \scalebox{0.72}{
    \begin{tabular}{|l|llll|l||llllllllll|l|}
        \hline
        \textbf{Fine-tuning} & \multicolumn{5}{c|}{\textbf{\color{blue}{Task 3}\color{black} : Open-I}} & \multicolumn{11}{c|}{\textbf{\color{red}{Task 4}\color{black} : MIMIC}}  \\
        
        \hline
       & \cellcolor[rgb]{ .886,  .937,  .855}\textbf{C1} & \cellcolor[rgb]{ .886,  .937,  .855}\textbf{C2} & \cellcolor[rgb]{ .886,  .937,  .855}\textbf{C3} & \cellcolor[rgb]{ .886,  .937,  .855}\textbf{C4} & \cellcolor[rgb]{ .706,  .776,  .906}\textbf{Overall} & \cellcolor[rgb]{ .886,  .937,  .855}\textbf{C1} & \cellcolor[rgb]{ .886,  .937,  .855}\textbf{C2} & \cellcolor[rgb]{ .886,  .937,  .855}\textbf{C3} & \cellcolor[rgb]{ .886,  .937,  .855}\textbf{C4} & \cellcolor[rgb]{ .886,  .937,  .855}\textbf{C5} & \cellcolor[rgb]{ .886,  .937,  .855}\textbf{C6} & \cellcolor[rgb]{ .886,  .937,  .855}\textbf{C7} & \cellcolor[rgb]{ .886,  .937,  .855}\textbf{C8} & \cellcolor[rgb]{ .886,  .937,  .855}\textbf{C9} & \cellcolor[rgb]{ .886,  .937,  .855}\textbf{C10} & \cellcolor[rgb]{ .706,  .776,  .906}\textbf{Overall}\\
        \hline \hline
        Full fine-tuning & 71.51 & 70.25 & 72.14 & 61.21 & 68.78 & 68.90 & 67.70 & 66.36 & 66.48 & 68.43 & 67.72 & 68.69 & 68.03 & 70.43 & 68.44 & 68.12\\ \hline
        Classifier only (LB) & 61.60 & 61.92 & 59.58 & 55.22 & 59.58 & 63.70 & 61.40 & 61.00 & 56.78 & 62.39 & 64.31 & 63.62 & 60.55 & 64.06 & 62.61 & 62.04 \\
        AdapterFusion & 67.74 & 65.83 & 66.74 & 54.48 & 63.70 & 67.33 & 64.65 & 63.24 & 60.49 & 66.05 & 67.05 & 66.53 & 66.41 & 64.72 & 68.39 & 65.49 \\
        Houlsby Adapter & 66.79  & 62.88 & 65.29 & 57.26 & 63.05 & 67.81 & 64.18 & 63.82 & 60.12 & 66.38 & 66.89 & 67.04 & 66.95 & 65.76 & 67.82 &  65.68 \\
        Parallel Adapter & 66.71 & 64.33 & 64.80 & 58.12 & 63.49 & 67.44 & 64.86 & 63.04 & 61.14 & 66.39 & 66.78 & 67.13 & 66.46 & 65.02 & 66.57 & 65.48  \\
        Compacter & 66.29 & 64.75 & 65.53 & 56.89 & 63.37 & 67.29 & 64.65 & 63.17 & 59.48 & 65.47 & 67.33 & 66.42 & 66.12 & 65.25 & 66.19 & 65.14  \\
        LayerNorm & 63.17 & 62.88 & 63.00 & 55.36 & 61.10 & 67.03 & 63.70 & 61.50 & 60.73 & 64.31 & 65.78 & 65.37 & 64.56 & 65.06 & 66.44 & 64.45  \\
        LoRA & 60.42 & 58.67 & 59.48 & 54.90 & 58.37 & 64.73 & 61.67 & 61.67 & 56.31 & 62.80 & 65.36 & 65.17 & 63.59 & 63.09 & 64.10 &  62.85 \\
        Bias & 63.53 & 62.63 & 62.81 & 56.90 & 61.47 & 65.96 & 63.09 & 61.99 & 59.85 & 63.76 & 65.53 & 63.57 & 63.91 & 65.21 & 66.38 & 64.12  \\
        PromptFL & 63.52 & 62.04 & 60.93 & 57.34 & 60.96 & 64.89 & 62.20 & 61.87 & 57.31 & 62.97 & 64.67 & 65.67 &  64.39 & 63.76 & 64.18 &  63.19 \\

        FedDAT (AdapterFusion) & 66.34 & 63.57 & 63.57 & 57.96 & 62.86 & 66.39 & 63.28 &62.20  & 60.34 & 66.94  & 67.32 & 67.34 & 67.76& 65.52 & 65.77 &  65.29 \\
        FedDAT (Houlsby) & 67.17 & 65.27 & 63.79 & 58.60 & 63.71 & 66.84 & 62.99 & 62.34 & 60.00 & 67.20 & 66.63 & 67.88 & 67.12 & 65.91 & 65.18 &  65.20 \\
        FedDAT (Parallel) & 65.40 & 64.89 & 63.34 & 55.55 & 62.30 & 66.10 & 63.10 & 61.27 & 61.39 & 65.83 & 67.35 & 66.36 & 66.76 &65.19  & 67.84 & 65.12  \\
        FedDAT (Compacter) & 67.55 & 66.19 & 64.27 & 58.06 & 64.02 & 67.09 & 62.56 & 61.80 & 59.03 & 64.68 & 66.79 & 67.24 & 67.40 & 65.39 & 66.02 &  64.80 \\
    
        FedPIA (Houlsby) \textbf{(ours)} & 70.99 & 69.76& \textbf{72.08} & 60.26 & 68.27 & \textbf{68.94} & \textbf{68.25} & \textbf{67.02} & 65.24 & \textbf{68.31} & \textbf{68.78} & \textbf{70.34} & 68.09 & 69.02 & \textbf{68.06} &  \textbf{68.21}  \\
        FedPIA (Parallel) \textbf{(ours)} & 70.20 & \textbf{70.14} & 71.78 & \textbf{61.01} & 68.28 & 68.76 & 66.12 & 65.44 & 65.67 & 67.30 & 67.42 & 68.18 & 67.46 & 69.75 & 67.96 &  67.41  \\
        FedPIA (Compacter) \textbf{(ours)} & \textbf{71.32} & 70.05 & 71.73 & 60.93 & \textbf{68.51} & 68.26 & 67.22 & 66.16 & \textbf{66.31} & 66.02 & 68.73& 68.91 &  \textbf{68.36}  & \textbf{69.89} & 67.02 &  67.69 \\
        \hline
    \end{tabular}}
    \vspace{-2.5mm}   
\end{table*}

\begin{table}[t]
    \centering
    \caption{Performance comparison of FedPIA with other methods on \color{green}{Task 5} \color{black}  (in terms of F1 score). V implies VQA}
    \vspace{-2mm}
    \scalebox{0.65}{
    \begin{tabular}{l|lllll|l}
        \hline
        \textbf{Fine-tuning}  & \cellcolor[rgb]{ .886,  .937,  .855}\textbf{Open-I} & \cellcolor[rgb]{ .886,  .937,  .855}\textbf{MIMIC} & \cellcolor[rgb]{ .886,  .937,  .855}\textbf{Slake} & \cellcolor[rgb]{ .886,  .937,  .855}\textbf{V-Med} & \cellcolor[rgb]{ .886,  .937,  .855}\textbf{V-Rad} & \cellcolor[rgb]{ .706,  .776,  .906}\textbf{Overall} \\
        \hline \hline
        Full fine-tuning & 74.22 & 65.34 & 97.78 & 97.44 & 98.76 & 86.71    \\  \hline
        Classifier only (LB) & 64.51 & 64.02 & 97.33 & 97.38 & 98.65 & 84.38  \\
        AdapterFusion & 72.27 & 63.41 & 97.76 & 97.60 & 98.71 & 85.95   \\
        Houlsby Adapter & 69.11 & 66.01 & 97.81 & 97.50 & 98.70 & 85.83   \\
        Parallel Adapter & 70.59 & 64.82 & 97.77 & 97.34 & 98.69 & 85.84  \\
        Compacter & 69.57 & 65.67 & 97.79 & 97.48 & 98.67 & 85.84  \\
        LayerNorm & 68.49 & 65.28 & 97.78 & 97.34 & 98.68 & 85.51  \\
        LoRA & 66.89 & 64.13 & 97.19 & 97.31 & 98.61 & 84.83  \\
        Bias & 67.99 & 64.81 & 97.67 & 97.45 & 98.69 & 85.32  \\
        PromptFL & 65.65 & 63.91 & 97.15 & 97.42 & 98.68 & 84.56 \\

        FedDAT (AdapterFusion) & 70.89 & 63.66 & 97.73 & 97.64 & 98.73 & 85.73 \\
        FedDAT (Houlsby) & 69.45 & 64.29 & 97.80 & 97.58 & 98.66 & 85.56 \\
        FedDAT (Parallel) & 70.88 & 65.03 &  97.82& 97.35 & 98.70 & 85.96  \\
        FedDAT (Compacter) & 70.19 & 64.86 & 97.75 & 97.54 & 98.60 &  85.79\\
    
        FedPIA (Houlsby) \textbf{(ours)} & \textbf{76.80} & \textbf{68.58} & \textbf{98.83}  & \textbf{98.73} & \textbf{99.64} & \textbf{88.52}  \\
        FedPIA (Parallel) \textbf{(ours)} & 75.68 & 67.01 & 98.44 & 97.96 & 98.95 & 87.61 \\
        FedPIA (Compacter) \textbf{(ours)} & 76.26 & 68.33 & 98.16 & 98.50 & 99.42 & 88.13  \\
        \hline
    \end{tabular}}
    \vspace{-2mm}   
\end{table}


    

\begin{table}[t]
    \centering
    \caption{Ablation study for all five tasks in terms of (overall): Accuracy for Tasks \color{magenta}1\color{black}, \color{cyan}2 \color{black} and F1 score for Tasks \color{blue}3\color{black}, \color{red}4\color{black}, and \color{green}5}
    \vspace{-2mm}
    \scalebox{0.8}{
    \begin{tabular}{l|lllll}
        \hline
        \textbf{Fine-tuning}  & \textbf{\color{magenta}{Task 1}\color{black} } & \textbf{\color{cyan}{Task 2}\color{black} } & \textbf{\color{blue}{Task 3}\color{black} } & \textbf{\color{red}{Task 4}\color{black} } & \textbf{\color{green}{Task 5}\color{black} }  \\
        \hline 
        \rowcolor[rgb]{ .988,  .894,  .839} & \multicolumn{5}{c}{\textbf{Backbone architecture: ViLT}}  \\
        \hline
        FedPIA  & 46.66 & 87.97 & 68.27  & 68.21 & 88.52 \\ \hline
   w/o any PIA  & 43.04 & 82.46 & 63.05 & 65.68 & 85.83  \\
   w/o server PIA  & 45.35 & 85.59 & 66.47 & 67.40 & 87.24  \\
   w/o client PIA  & 44.47 & 85.02 & 65.58 & 66.28 & 86.97 \\ 
   w/ weight-based PIA  & 46.02 &87.13 & 67.62 & 67.87 & 87.93 \\ \hline

       \rowcolor[rgb]{ .988,  .894,  .839} & \multicolumn{5}{c}{\textbf{Backbone architecture: ALBEF}}  \\ 
        \hline
        FedPIA  & 48.24 & 89.05 & 68.78 & 69.46 & 84.49 \\ \hline
   w/o any PIA & 43.17 & 83.37 & 62.76 &   65.43 & 80.28\\
   w/o server PIA  & 46.58 & 86.44 & 65.92 & 67.88 & 83.16  \\
   w/o client PIA  & 45.20 & 85.29 & 64.87 & 66.38 &  82.00\\
w/ weight-based PIA  & 47.75 & 88.38 & 68.09 & 68.65 &  83.99\\

        \hline
    \end{tabular}}
    \vspace{-2mm}   
\end{table}
\section{Experiments and Results}

\subsection{Tasks and Datasets}
We assess the performance of our proposed method with two prominent Vision Language foundation models, \textit{viz}., ViLT and ALBEF, and for three FL task settings: (a) Visual Question Answering, (b) Image and Text-based Disease Classification, (c) Heterogeneous tasks combining both (a) and (b). In order to ensure the real-world applicability of FedPIA, we conduct experiments on multiple well-known and challenging medical datasets as discussed below. 

\subsubsection{(a) Visual Question Answering:} We consider \textbf{two} scenarios with data of varying sizes, class counts, and complexity:

\noindent
\begin{itemize}
    \item[(i)] \textbf{\color{magenta}{Task 1}\color{black} : Five-client setting} with SLAKE \cite{liu2021slake}, VQA-RAD \cite{lau2018dataset}, VQA-Med 2019 \cite{ben2019vqa}, VQA-Med 2020 \cite{BenAbacha2020VQAMed}, and VQA-Med 2021 \cite{ben2021overview}.
    
    \item[(ii)] \textbf{\color{cyan}{\color{cyan}{Task 2}\color{black} }\color{black}: Modality specific Eight-client setting} leveraging \cite{Hu_2024_CVPR} where \textbf{Client 1 (CT)} includes 3 CT datasets, \textbf{Client 2 (US)} includes Ultrasound images, \textbf{Client 3 (OCT)} includes 2 datasets, \textbf{Client 4 (Fundus images)} includes 8 fundus datasets, \textbf{Client 5 (Microscopy)} includes 5 datasets, \textbf{Client 6 (Histopathology)} includes 4 datasets, \textbf{Client 7 (Dermatoscopy)} includes 7 different skin datasets, \textbf{Client 8 (X-Ray)} includes 11 X-Ray datasets. See \textbf{Suppl. \S B} for details.
    \end{itemize}
\subsubsection{(b) Image and text-based disease classification:} Following \cite{saha2024examiningmodalityincongruitymultimodal}, we consider \textbf{two} heterogeneous FL settings for Chest X-Ray and Radiology report-based multi-label disease detection: 

\noindent
(i) \textbf{\color{blue}{Task 3}\color{black} :} 4 client-scenario using Open-I dataset, 

\noindent
(ii) \textbf{\color{red}{Task 4}\color{black} :} 10 client scenario with MIMIC dataset. 

Following standard procedures, we use Dirichlet distributions with $\gamma=0.5$ to simulate non-IID client data partitions from each dataset. There are 15 disease classes in each of the datasets, \textit{viz.}, Support Device, Pleural Effusion, Consolidation, Pneumothorax, Lung Opacity, Enlarged Cardiomediastinum, Atelectasis, Others, Cardiomegaly, Lung lesion, Edema, Fracture, Pneumonia, Pleural other, and No finding.

\subsubsection{(c) Heterogeneous tasks:} We consider a \textbf{task-heterogeneous} setting for \textbf{\color{green}{Task 5}\color{black} } combining three Visual Question answering clients, \textit{viz}., SLAKE, VQA-RAD, VQA-Med 2019, and two disease-classification clients, \textit{viz}., Open-I and MIMIC.
\vspace{-1mm}
\subsection{Training and Implementation Details}

To demonstrate the effectiveness of FedPIA across various VLM models, we adopt two types of VLM transformer architectures: (a) encoder-only backbone \textit{i.e.}, ViLT \cite{kim2021vilt}, and (b) encoder-decoder backbone \textit{i.e.}, ALBEF \cite{li2021align}. We fix the initial learning rate $\eta = 0.0001$ and batch size $B = 16$. We use the AdamW optimizer and a learning rate scheduler with linear decay following \cite{chen2024feddat}. We also use a weight decay of 0.01 with a total of 30 communication rounds for federated fine-tuning including $10\%$ warmup rounds \cite{chen2024feddat}. Each client has task-specific linear classification heads. 

\subsubsection{Baselines:} Our baselines are: 1) Full fine-tuning, 2) Local classifier fine-tuning, 3) AdapterFusion \cite{pfeiffer2020adapterfusion}, 4) Houlsby Adapter \cite{houlsby2019parameter}, 5) Parallel Adapter \cite{he2022towards}, 6) Compacter \cite{karimi2021compacter}, 7) LayerNorm \cite{basu2023strong}, 8) LoRA \cite{hu2022lora}, 9) Bias tuning \cite{tinyTL}, 10) PromptFL \cite{guo2023promptfl}, and 11-14) FedDAT \cite{chen2024feddat} with 4 adapters variants. Note that `Local classifier' refers to client-specific training of local classifier heads without any federated learning and hence it is considered as the Lower Bound (LB). See \textbf{Suppl. \S B} for more details.

\vspace{-1mm}
\subsection{Results on Visual Question Answering (Tasks \color{magenta}1\color{black}, \color{cyan}2\color{black})}
Tables 1 and 2 show the performance comparison of our models and the baselines for Tasks \color{magenta}1 \color{black} and \color{cyan}2 \color{black} respectively. For \color{magenta}Task 1\color{black}, we show the accuracy of the models in answering open-ended and closed questions separately in each dataset except VQA-Med 2021 which does not possess any closed questions. We observe that FedPIA outperforms all naive PEFT-FL and SOTA methods for all VQA tasks and scenarios. The performance of adapter-based PEFT baselinesdegrade under heterogeneity conditions, as visualized in the loss curve from Fig. 3 (a). This is mainly due to the baselines failing to properly integrate client-agnostic knowledge with client-specific knowledge from multiple adapters trained on diverse data, modalities, and tasks. On the contrary, FedPIA shows robust and consistent performance irrespective of data and modality heterogeneity, which in turn, demonstrates that our permutation and integration mechanism effectively handles the challenging non-IID scenario by bridging the gap between adapters in weight space. Our method achieves an overall mean improvement of $\mathbf{3.89\%}$ and {$\mathbf{5.18\%}$} for Tasks \color{magenta}1 \color{black} and \color{cyan}2 \color{black} respectively over FedDAT across all adapter configurations. In VQA-Med 2021 (from Tab. 1), FedDAT performs better than full fine-tuning. This is possibly because adapters in FedDAT are well-suited to retain task-specific adaptations and hence perform better than full fine-tuning which spreads updates across all parameters, diluting task-specific information. For further experiments or analysis, see \textbf{Suppl.\S C}

\begin{figure}[htbp]
\vspace{-2mm}
    \centering
\includegraphics[width=1\columnwidth]{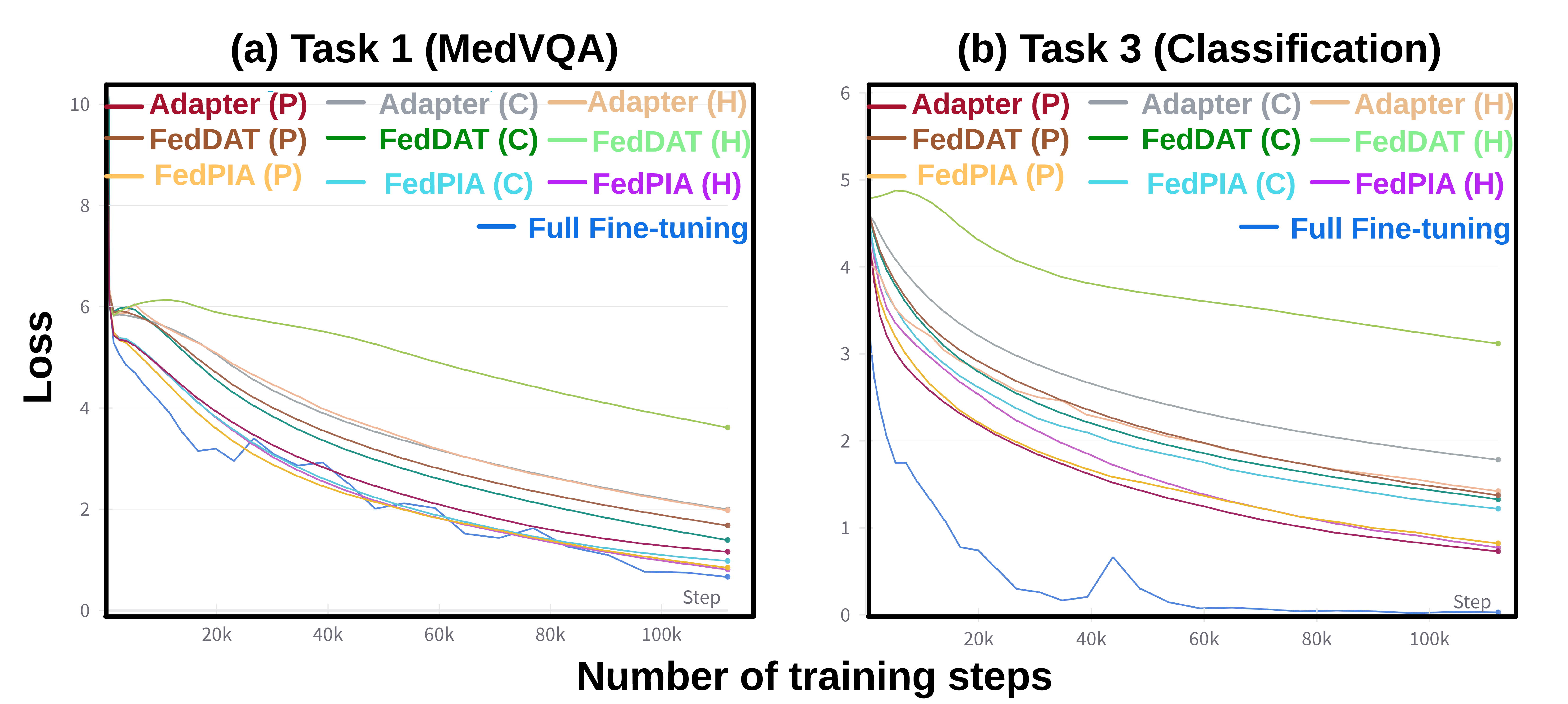}
    \caption{Convergence analysis for Tasks \color{magenta}1 \color{black} \& \color{blue}3 \color{black}. C, P, and H refer to Compacter, Parallel adapter, and Houlsby adapter.}
\label{fig1}
\end{figure}

\begin{figure}[htbp]
\vspace{-2mm}
    \centering
\includegraphics[width=1\columnwidth]{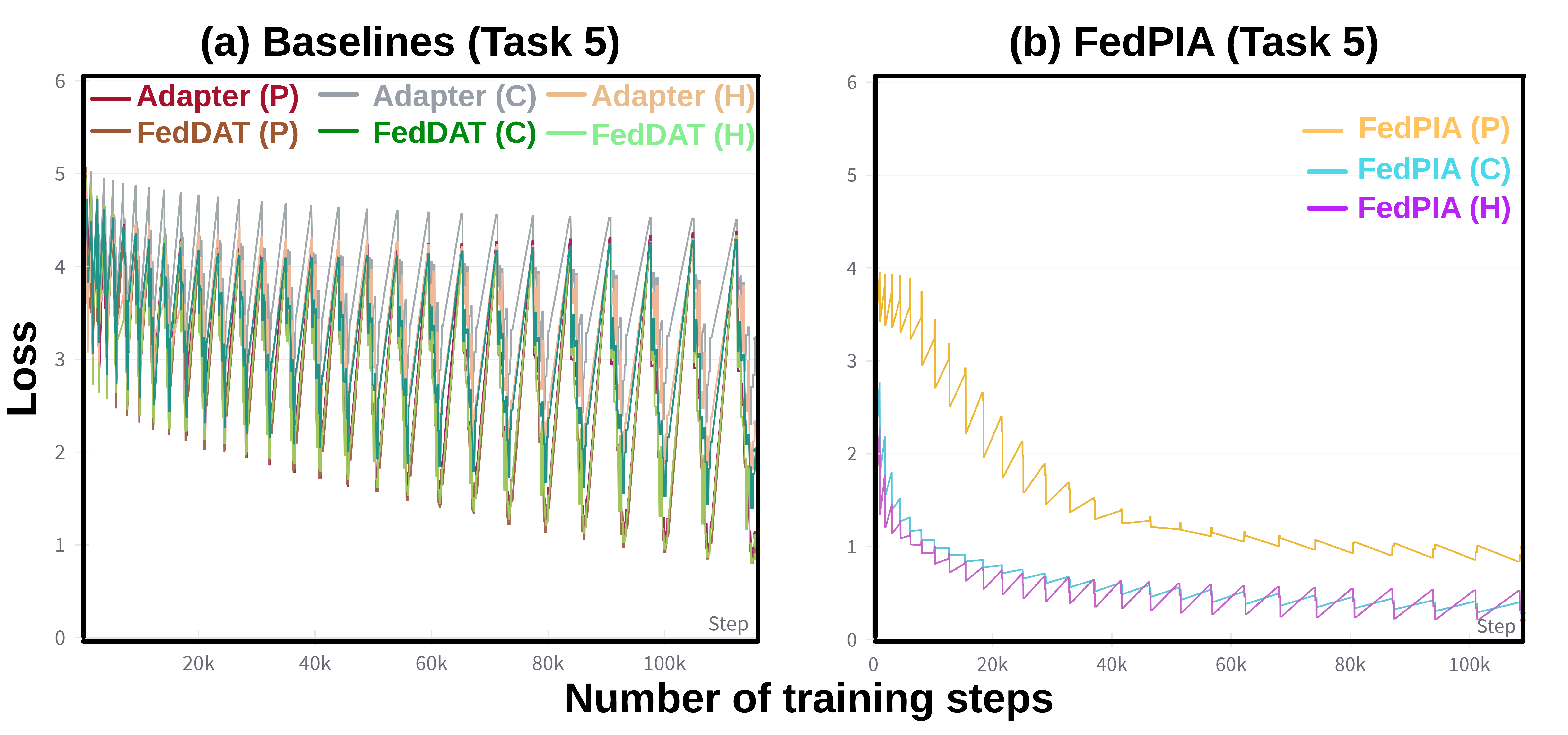}
    \caption{Convergence analysis of adapter-based baseline models (left) and FedPIA (right) for \color{green}{Task 5} \color{black}  (Heterogeneous task). The peaks and troughs represent the losses at the start and end of each communicating rounds. The reduction in magnitude of spikes (right) show that FedPIA bridges the gap between global and local adapters leading to a faster, stable convergence with less oscillation.}
\label{fig1}
\vspace{-4mm}
\end{figure}

\subsection{Results on Multilabel Disease Detection (Tasks \color{blue}3\color{black}, \color{red}4\color{black})}
Table 3 shows that our method demonstrates consistent performance improvement with respect to the baselines in each client for the multi-label classification task on both the datasets. We notice that the baseline methods show deterioration in performance due to the statistical heterogeneity introduced by inter-client class distribution shift. FedPIA outperforms FedDAT approximately by $\mathbf{5.1\%}$ and $\mathbf{2.73\%}$ in F1-score for Open-I and MIMIC datasets across all clients and over all adapter configurations. These results show that our method more effectively utilizes the knowledge from multiple adapters in statistically heterogeneous FL settings. This is further supported by the visualization of convergence analysis in Fig. 3 (b) where FedPIA loss curves are seen to be closest to the full fine-tuning loss curve.
\subsection{Results on Heterogeneous Task (\color{green}{Task 5}\color{black} )}
Table 4 reports the performance of FedPIA and baselines under task-heterogeneous settings with VQA and classification datasets simultaneously. Fig. 4 further visualizes the corresponding loss curves for convergence analysis. It demonstrates that FedPIA results in much faster and more stable convergence that the adapter-based baselines by reducing the oscillations resulting from task heterogeneity. This reduction is achieved by the two-fold alignment of local and global adapters in clients and server. Interestingly, our model not only outperforms the baselines in each client, but also full fine-tuning (on average by $\mathbf{1.81}\%$) which suggests that our method more effectively preserves learned knowledge from diverse clients even with a small number of parameters whereas the baselines and full fine-tuning suffer from catastrophic forgetting due to knowledge interference from heterogeneous tasks and data distribution. The substantial boost in Open-I client model is attributed to the significantly larger overall FL dataset size in Task 5 (34323 samples across 5 clients) compared to Task 3 (2837 samples across 4 clients). However, MIMIC client model in Task 4 is developed using 73348 samples across 10 clients and hence, it does not show a notable learning advantage in Task 5.
\subsection{Ablation Studies}
We study the impact of each component of FedPIA (Houlsby) via ablation analysis in Table 5. Our model with either client-based or server-based PIA alone is observed to outperform the baselines. Removing either of client-based and server-based adaptation leads to a drop in performance. Greater performance degradation is observed when dropping client-level PIA, which suggests that client-based PIA captures more essential information from the permuted global adapter. We also show that replacing the activation by weight-based ground cost computation in the clients leads to slight decrease in adapter performance.

To investigate the impact of different hyperparameters and client size, we vary the learning rate and batch size, as well as progressively reduce the dataset size in each client from $100\%$ in steps of $20\%$. We also investigate the scalability of FedPIA by increasing the number of clients. \textbf{See Suppl. \S C}. 
\section{Conclusion}
The main contributions of the work are as follows:
\begin{enumerate}
    \item  We studied the practical problem of parameter-efficient fine-tuning of foundation models in multimodal FL for tackling data- and resource-constraints. We analysed different real-world problem scenarios with the overall goal of performing medical visual question answering or vision-language-based disease classification or both simultaneously. Through five different tasks, we, for the first time, investigated three FL heterogeneity settings - statistical heterogeneity, modality heterogeneity, as well as task heterogeneity in the context of PEFT-FL.

    \item We proposed a novel method, Federated Learning via Permutation and Integration of Adapters, that exploits Wasserstein Barycenters for shuffling and combining adapters. We demonstrated this to be particularly effective in bringing adapters closer in data- and task-heterogeneous situations where the adapters are distant in parameter space. Our method does not require retraining for alignment or further knowledge distillation like existing methods, thereby adding no training overhead. Besides, FedPIA is orthogonal to existing FL aggregation schemes and can be used in conjunction with those. 
    \item For evaluating the performance of FedPIA under modality heterogeneity, we developed a modality-specific FL setup using 41 medical image datasets. Through comprehensive experiments, we showed that FedPIA outperforms both SOTA and naive combinations of PEFT and FL. The results demonstrate that our proposed method can achieve, and even surpass, the performance of fully fine-tuned methods across diverse tasks, for fine-tuning VLMs in heterogeneous FL.
\end{enumerate}

\section{Acknowledgments}
This work was supported in part by the UK EPSRC (Engineering and Physical Research Council) Programme Grant EP/T028572/1 (VisualAI), a  UK EPSRC Doctoral Training Partnership award, the UKRI grant EP/X040186/1 (Turing AI Fellowship), and the InnoHK-funded Hong Kong Centre for Cerebro-cardiovascular Health Engineering (COCHE) Project 2.1 (Cardiovascular risks in early life and fetal echocardiography). FW is supported by the EPSRC Centre for Doctoral Training in Health Data Science (EP/S02428X/1), by the Anglo-Austrian Society, and by an Oxford-Reuben scholarship. PS acknowledges Yash Bhalgat for the insightful discussions and valuable inputs.



\bibliography{aaai25}

\begin{thebibliography}{54}
\providecommand{\natexlab}[1]{#1}

\bibitem[{Abacha et~al.(2020)Abacha, Datla, Hasan, Demner-Fushman, and M{\"u}ller}]{BenAbacha2020VQAMed}
Abacha, A. S.~B.; Datla, V.~V.; Hasan, S.~A.; Demner-Fushman, D.; and M{\"u}ller, H. 2020.
\newblock Overview of the VQA-Med Task at ImageCLEF 2020: Visual Question Answering and Generation in the Medical Domain.
\newblock In \emph{CLEF 2020 Working Notes}, volume 2696 of \emph{CEUR Workshop Proceedings}.

\bibitem[{Acar et~al.(2021)Acar, Zhao, Navarro, Mattina, Whatmough, and Saligrama}]{acar2021federated}
Acar, D. A.~E.; Zhao, Y.; Navarro, R.~M.; Mattina, M.; Whatmough, P.~N.; and Saligrama, V. 2021.
\newblock Federated learning based on dynamic regularization.
\newblock \emph{arXiv preprint arXiv:2111.04263}.

\bibitem[{Akash, Li, and Trillos(2022)}]{akash2022wasserstein}
Akash, A.~K.; Li, S.; and Trillos, N.~G. 2022.
\newblock Wasserstein barycenter-based model fusion and linear mode connectivity of neural networks.
\newblock \emph{arXiv preprint arXiv:2210.06671}.

\bibitem[{Antol et~al.(2015)Antol, Agrawal, Lu, Mitchell, Batra, Zitnick, and Parikh}]{antol2015vqa}
Antol, S.; Agrawal, A.; Lu, J.; Mitchell, M.; Batra, D.; Zitnick, C.~L.; and Parikh, D. 2015.
\newblock Vqa: Visual question answering.
\newblock In \emph{Proceedings of the IEEE international conference on computer vision}.

\bibitem[{Basu et~al.(2023)Basu, Massiceti, Hu, and Feizi}]{basu2023strong}
Basu, S.; Massiceti, D.; Hu, S.~X.; and Feizi, S. 2023.
\newblock Strong Baselines for Parameter Efficient Few-Shot Fine-tuning.
\newblock \emph{arXiv preprint arXiv:2304.01917}.

\bibitem[{Ben~Abacha et~al.(2019)Ben~Abacha, Hasan, Datla, Demner-Fushman, and M{\"u}ller}]{ben2019vqa}
Ben~Abacha, A.; Hasan, S.~A.; Datla, V.~V.; Demner-Fushman, D.; and M{\"u}ller, H. 2019.
\newblock Vqa-med: Overview of the medical visual question answering task at imageclef 2019.
\newblock In \emph{Proceedings of CLEF (Conference and Labs of the Evaluation Forum) 2019 Working Notes}.

\bibitem[{Ben~Abacha et~al.(2021)Ben~Abacha, Sarrouti, Demner-Fushman, Hasan, and M{\"u}ller}]{ben2021overview}
Ben~Abacha, A.; Sarrouti, M.; Demner-Fushman, D.; Hasan, S.~A.; and M{\"u}ller, H. 2021.
\newblock Overview of the vqa-med task at imageclef 2021: Visual question answering and generation in the medical domain.
\newblock In \emph{Proceedings of the CLEF 2021 Conference and Labs of the Evaluation Forum-working notes}.

\bibitem[{Ben~Zaken, Goldberg, and Ravfogel(2022)}]{ben-zaken-etal-2022-bitfit}
Ben~Zaken, E.; Goldberg, Y.; and Ravfogel, S. 2022.
\newblock {B}it{F}it: Simple Parameter-efficient Fine-tuning for Transformer-based Masked Language-models.
\newblock In \emph{Proceedings of the Association for Computational Linguistics (Volume 2: Short Papers)}, 1--9. Dublin, Ireland.

\bibitem[{Cai et~al.(2020)Cai, Gan, Zhu, and Han}]{tinyTL}
Cai, H.; Gan, C.; Zhu, L.; and Han, S. 2020.
\newblock TinyTL: Reduce Memory, Not Parameters for Efficient On-Device Learning.
\newblock In \emph{Advances in Neural Information Processing Systems}, volume~33, 11285--11297.

\bibitem[{Chen et~al.(2024)Chen, Zhang, Krompass, Gu, and Tresp}]{chen2024feddat}
Chen, H.; Zhang, Y.; Krompass, D.; Gu, J.; and Tresp, V. 2024.
\newblock Feddat: An approach for foundation model finetuning in multi-modal heterogeneous federated learning.
\newblock In \emph{Proceedings of the AAAI Conference on Artificial Intelligence}, volume~38, 11285--11293.

\bibitem[{Chen et~al.(2022)Chen, Xu, Guo, Wang, Zhang, and Wang}]{chen2022fedtune}
Chen, J.; Xu, W.; Guo, S.; Wang, J.; Zhang, J.; and Wang, H. 2022.
\newblock Fedtune: A deep dive into efficient federated fine-tuning with pre-trained transformers.
\newblock \emph{arXiv preprint arXiv:2211.08025}.

\bibitem[{Frankle, Schwab, and Morcos(2021)}]{frankle2021trainingbatchnormbatchnormexpressive}
Frankle, J.; Schwab, D.~J.; and Morcos, A.~S. 2021.
\newblock Training BatchNorm and Only BatchNorm: On the Expressive Power of Random Features in CNNs.
\newblock arXiv:2003.00152.

\bibitem[{Guo, Guo, and Wang(2023)}]{guo2023pfedprompt}
Guo, T.; Guo, S.; and Wang, J. 2023.
\newblock Pfedprompt: Learning personalized prompt for vision-language models in federated learning.
\newblock In \emph{Proceedings of the ACM Web Conference 2023}, 1364--1374.

\bibitem[{Guo et~al.(2023)Guo, Guo, Wang, Tang, and Xu}]{guo2023promptfl}
Guo, T.; Guo, S.; Wang, J.; Tang, X.; and Xu, W. 2023.
\newblock Promptfl: Let federated participants cooperatively learn prompts instead of models-federated learning in age of foundation model.
\newblock \emph{IEEE Transactions on Mobile Computing}.

\bibitem[{He et~al.(2022)He, Zhou, Ma, Berg-Kirkpatrick, and Neubig}]{he2022towards}
He, J.; Zhou, C.; Ma, X.; Berg-Kirkpatrick, T.; and Neubig, G. 2022.
\newblock Towards a Unified View of Parameter-Efficient Transfer Learning.
\newblock In \emph{International Conference on Learning Representations}.

\bibitem[{Hernandez-Cruz et~al.(2024)Hernandez-Cruz, Saha, Sarker, and Noble}]{bdcc8090099}
Hernandez-Cruz, N.; Saha, P.; Sarker, M. M.~K.; and Noble, J.~A. 2024.
\newblock Review of Federated Learning and Machine Learning-Based Methods for Medical Image Analysis.
\newblock \emph{Big Data and Cognitive Computing}, 8(9).

\bibitem[{Houlsby et~al.(2019)Houlsby, Giurgiu, Jastrzebski, Morrone, De~Laroussilhe, Gesmundo, Attariyan, and Gelly}]{houlsby2019parameter}
Houlsby, N.; Giurgiu, A.; Jastrzebski, S.; Morrone, B.; De~Laroussilhe, Q.; Gesmundo, A.; Attariyan, M.; and Gelly, S. 2019.
\newblock Parameter-efficient transfer learning for NLP.
\newblock In \emph{International conference on machine learning}, 2790--2799. PMLR.

\bibitem[{Hu et~al.(2022)Hu, yelong shen, Wallis, Allen-Zhu, Li, Wang, Wang, and Chen}]{hu2022lora}
Hu, E.~J.; yelong shen; Wallis, P.; Allen-Zhu, Z.; Li, Y.; Wang, S.; Wang, L.; and Chen, W. 2022.
\newblock Lo{RA}: Low-Rank Adaptation of Large Language Models.
\newblock In \emph{International Conference on Learning Representations}.

\bibitem[{Hu et~al.(2024)Hu, Li, Lu, Shao, He, Qiao, and Luo}]{Hu_2024_CVPR}
Hu, Y.; Li, T.; Lu, Q.; Shao, W.; He, J.; Qiao, Y.; and Luo, P. 2024.
\newblock OmniMedVQA: A New Large-Scale Comprehensive Evaluation Benchmark for Medical LVLM.
\newblock In \emph{Proceedings of the IEEE/CVF Conference on Computer Vision and Pattern Recognition (CVPR)}, 22170--22183.

\bibitem[{Jia et~al.(2022)Jia, Tang, Chen, Cardie, Belongie, Hariharan, and Lim}]{VPT_jia}
Jia, M.; Tang, L.; Chen, B.-C.; Cardie, C.; Belongie, S.; Hariharan, B.; and Lim, S.-N. 2022.
\newblock Visual Prompt Tuning.
\newblock In \emph{Computer Vision – ECCV 2022: 17th European Conference, Tel Aviv, Israel, October 23–27, 2022, Proceedings, Part XXXIII}, 709–727. Springer-Verlag.

\bibitem[{Karimi~Mahabadi, Henderson, and Ruder(2021)}]{karimi2021compacter}
Karimi~Mahabadi, R.; Henderson, J.; and Ruder, S. 2021.
\newblock Compacter: Efficient low-rank hypercomplex adapter layers.
\newblock \emph{Advances in Neural Information Processing Systems}, 34: 1022--1035.

\bibitem[{Karimireddy et~al.(2020)Karimireddy, Kale, Mohri, Reddi, Stich, and Suresh}]{karimireddy2020scaffold}
Karimireddy, S.~P.; Kale, S.; Mohri, M.; Reddi, S.; Stich, S.; and Suresh, A.~T. 2020.
\newblock Scaffold: Stochastic controlled averaging for federated learning.
\newblock In \emph{International conference on machine learning}. PMLR.

\bibitem[{Kim, Son, and Kim(2021)}]{kim2021vilt}
Kim, W.; Son, B.; and Kim, I. 2021.
\newblock Vilt: Vision-and-language transformer without convolution or region supervision.
\newblock In \emph{International conference on machine learning}, 5583--5594. PMLR.

\bibitem[{Lau et~al.(2018)Lau, Gayen, Ben~Abacha, and Demner-Fushman}]{lau2018dataset}
Lau, J.~J.; Gayen, S.; Ben~Abacha, A.; and Demner-Fushman, D. 2018.
\newblock A dataset of clinically generated visual questions and answers about radiology images.
\newblock \emph{Scientific data}, 5(1): 1--10.

\bibitem[{Lester, Al-Rfou, and Constant(2021)}]{lester2021power}
Lester, B.; Al-Rfou, R.; and Constant, N. 2021.
\newblock The power of scale for parameter-efficient prompt tuning.
\newblock \emph{arXiv preprint arXiv:2104.08691}.

\bibitem[{Li et~al.(2023)Li, Wu, Sun, Shen, Wu, and Tao}]{li2023visual}
Li, G.; Wu, W.; Sun, Y.; Shen, L.; Wu, B.; and Tao, D. 2023.
\newblock Visual prompt based personalized federated learning.
\newblock \emph{arXiv preprint arXiv:2303.08678}.

\bibitem[{Li et~al.(2021)Li, Selvaraju, Gotmare, Joty, Xiong, and Hoi}]{NEURIPS2021_50525975}
Li, J.; Selvaraju, R.; Gotmare, A.; Joty, S.; Xiong, C.; and Hoi, S. C.~H. 2021.
\newblock Align before Fuse: Vision and Language Representation Learning with Momentum Distillation.
\newblock In Ranzato, M.; Beygelzimer, A.; Dauphin, Y.; Liang, P.; and Vaughan, J.~W., eds., \emph{Advances in Neural Information Processing Systems}, volume~34, 9694--9705.

\bibitem[{Li, He, and Song(2021)}]{li2021model}
Li, Q.; He, B.; and Song, D. 2021.
\newblock Model-contrastive federated learning.
\newblock In \emph{Proceedings of the IEEE/CVF conference on computer vision and pattern recognition}, 10713--10722.

\bibitem[{Li et~al.(2020)Li, Sahu, Talwalkar, and Smith}]{li2020federated}
Li, T.; Sahu, A.~K.; Talwalkar, A.; and Smith, V. 2020.
\newblock Federated learning: Challenges, methods, and future directions.
\newblock \emph{IEEE signal processing magazine}, 37(3): 50--60.

\bibitem[{Li, Liu, and Bilen(2022)}]{li2022cross}
Li, W.-H.; Liu, X.; and Bilen, H. 2022.
\newblock Cross-domain few-shot learning with task-specific adapters.
\newblock In \emph{Proceedings of the IEEE/CVF Conference on Computer Vision and Pattern Recognition}, 7161--7170.

\bibitem[{Li and Liang(2021)}]{li2021prefix}
Li, X.~L.; and Liang, P. 2021.
\newblock Prefix-tuning: Optimizing continuous prompts for generation.
\newblock \emph{arXiv preprint arXiv:2101.00190}.

\bibitem[{Lian et~al.(2022)Lian, Zhou, Feng, and Wang}]{lian2022scaling}
Lian, D.; Zhou, D.; Feng, J.; and Wang, X. 2022.
\newblock Scaling \& shifting your features: A new baseline for efficient model tuning.
\newblock \emph{arXiv preprint arXiv:2210.08823}.

\bibitem[{Liu et~al.(2021)Liu, Zhan, Xu, Ma, Yang, and Wu}]{liu2021slake}
Liu, B.; Zhan, L.-M.; Xu, L.; Ma, L.; Yang, Y.; and Wu, X.-M. 2021.
\newblock Slake: A semantically-labeled knowledge-enhanced dataset for medical visual question answering.
\newblock In \emph{2021 IEEE 18th International Symposium on Biomedical Imaging (ISBI)}, 1650--1654. IEEE.

\bibitem[{Lu et~al.(2023)Lu, Hu, Wang, and Xie}]{lu2023fedclip}
Lu, W.; Hu, X.; Wang, J.; and Xie, X. 2023.
\newblock Fedclip: Fast generalization and personalization for clip in federated learning.
\newblock \emph{arXiv preprint arXiv:2302.13485}.

\bibitem[{McMahan et~al.(2017)McMahan, Moore, Ramage, Hampson, and y~Arcas}]{mcmahan2017communication}
McMahan, B.; Moore, E.; Ramage, D.; Hampson, S.; and y~Arcas, B.~A. 2017.
\newblock Communication-efficient learning of deep networks from decentralized data.
\newblock In \emph{Artificial intelligence and statistics}. PMLR.

\bibitem[{Nguyen, Munoz, and Jannesari(2024)}]{nguyen2024flora}
Nguyen, D.~P.; Munoz, J.~P.; and Jannesari, A. 2024.
\newblock Flora: Enhancing vision-language models with parameter-efficient federated learning.
\newblock \emph{arXiv preprint arXiv:2404.15182}.

\bibitem[{Pfeiffer et~al.(2020)Pfeiffer, Kamath, R{\"u}ckl{\'e}, Cho, and Gurevych}]{pfeiffer2020adapterfusion}
Pfeiffer, J.; Kamath, A.; R{\"u}ckl{\'e}, A.; Cho, K.; and Gurevych, I. 2020.
\newblock Adapterfusion: Non-destructive task composition for transfer learning.
\newblock \emph{arXiv preprint arXiv:2005.00247}.

\bibitem[{Rebuffi, Bilen, and Vedaldi(2018)}]{rebuffi2018efficient}
Rebuffi, S.-A.; Bilen, H.; and Vedaldi, A. 2018.
\newblock Efficient parametrization of multi-domain deep neural networks.
\newblock In \emph{Proceedings of the IEEE Conference on Computer Vision and Pattern Recognition}.

\bibitem[{Saha, Mishra, and Noble(2023)}]{saha2023rethinking}
Saha, P.; Mishra, D.; and Noble, J.~A. 2023.
\newblock Rethinking Semi-Supervised Federated Learning: How to co-train fully-labeled and fully-unlabeled client imaging data.
\newblock In \emph{International Conference on Medical Image Computing and Computer-Assisted Intervention}, 414--424. Springer.

\bibitem[{Saha et~al.(2024{\natexlab{a}})Saha, Mishra, Wagner, Kamnitsas, and Noble}]{saha2024examiningmodalityincongruitymultimodal}
Saha, P.; Mishra, D.; Wagner, F.; Kamnitsas, K.; and Noble, J.~A. 2024{\natexlab{a}}.
\newblock Examining Modality Incongruity in Multimodal Federated Learning for Medical Vision and Language-based Disease Detection.
\newblock arXiv:2402.05294.

\bibitem[{Saha et~al.(2024{\natexlab{b}})Saha, Wagner, Mishra, Peng, Thakur, Clifton, Kamnitsas, and Noble}]{saha2024f3ocusfederatedfinetuning}
Saha, P.; Wagner, F.; Mishra, D.; Peng, C.; Thakur, A.; Clifton, D.; Kamnitsas, K.; and Noble, J.~A. 2024{\natexlab{b}}.
\newblock F$^3$OCUS -- Federated Finetuning of Vision-Language Foundation Models with Optimal Client Layer Updating Strategy via Multi-objective Meta-Heuristics.
\newblock arXiv:2411.11912.

\bibitem[{Singh and Jaggi(2020)}]{singh2020model}
Singh, S.~P.; and Jaggi, M. 2020.
\newblock Model fusion via optimal transport.
\newblock \emph{Advances in Neural Information Processing Systems}, 33: 22045--22055.

\bibitem[{Su et~al.(2022)Su, Yang, Li, and Xue}]{su2022cross}
Su, S.; Yang, M.; Li, B.; and Xue, X. 2022.
\newblock Cross-domain federated adaptive prompt tuning for clip.
\newblock \emph{arXiv preprint arXiv:2211.07864}, 3.

\bibitem[{Sun et~al.(2022)Sun, Mendieta, Yang, and Chen}]{sun2022exploring}
Sun, G.; Mendieta, M.; Yang, T.; and Chen, C. 2022.
\newblock Exploring parameter-efficient fine-tuning for improving communication efficiency in federated learning.

\bibitem[{Touvron et~al.(2022)Touvron, Cord, El-Nouby, Verbeek, and J{\'e}gou}]{touvron2022three}
Touvron, H.; Cord, M.; El-Nouby, A.; Verbeek, J.; and J{\'e}gou, H. 2022.
\newblock Three things everyone should know about vision transformers.
\newblock In \emph{Computer Vision--ECCV 2022: 17th European Conference, Tel Aviv, Israel, October 23--27, 2022, Proceedings, Part XXIV}. Springer.

\bibitem[{Wagner et~al.(2023)Wagner, Li, Saha, and Kamnitsas}]{wagner2023post}
Wagner, F.; Li, Z.; Saha, P.; and Kamnitsas, K. 2023.
\newblock Post-Deployment Adaptation with Access to Source Data via Federated Learning and Source-Target Remote Gradient Alignment.
\newblock In \emph{International Workshop on Machine Learning in Medical Imaging}, 253--263. Springer.

\bibitem[{Wagner et~al.(2024)Wagner, Xu, Saha, Liang, Whitehouse, Menon, Newcombe, Voets, Noble, and Kamnitsas}]{wagner2024feasibilityfederatedlearningclient}
Wagner, F.; Xu, W.; Saha, P.; Liang, Z.; Whitehouse, D.; Menon, D.; Newcombe, V.; Voets, N.; Noble, J.~A.; and Kamnitsas, K. 2024.
\newblock Feasibility of Federated Learning from Client Databases with Different Brain Diseases and MRI Modalities.
\newblock arXiv:2406.11636.

\bibitem[{Yang et~al.(2024)Yang, Su, Li, and Xue}]{yang2024exploring}
Yang, M.; Su, S.; Li, B.; and Xue, X. 2024.
\newblock Exploring One-Shot Semi-supervised Federated Learning with Pre-trained Diffusion Models.
\newblock In \emph{Proceedings of the AAAI Conference on Artificial Intelligence}, volume~38.

\bibitem[{Yu et~al.(2023)Yu, Liu, Wang, Xu, and Liu}]{yu2023multimodal}
Yu, Q.; Liu, Y.; Wang, Y.; Xu, K.; and Liu, J. 2023.
\newblock Multimodal federated learning via contrastive representation ensemble.
\newblock \emph{arXiv preprint arXiv:2302.08888}.

\bibitem[{Yu, Mu{\~n}oz, and Jannesari(2023)}]{yu2023federated}
Yu, S.; Mu{\~n}oz, J.~P.; and Jannesari, A. 2023.
\newblock Federated foundation models: Privacy-preserving and collaborative learning for large models.
\newblock \emph{arXiv preprint arXiv:2305.11414}.

\bibitem[{Zellers et~al.(2019)Zellers, Bisk, Farhadi, and Choi}]{zellers2019recognitioncognitionvisualcommonsense}
Zellers, R.; Bisk, Y.; Farhadi, A.; and Choi, Y. 2019.
\newblock From Recognition to Cognition: Visual Commonsense Reasoning.
\newblock arXiv:1811.10830.

\bibitem[{Zeng, Yue, and Wang(2024)}]{zeng2024open}
Zeng, H.; Yue, Z.; and Wang, D. 2024.
\newblock Open-Vocabulary Federated Learning with Multimodal Prototyping.
\newblock \emph{arXiv preprint arXiv:2404.01232}.

\bibitem[{Zhang et~al.(2024)Zhang, Vahidian, Kuo, Li, Zhang, Yu, Wang, and Chen}]{zhang2024towards}
Zhang, J.; Vahidian, S.; Kuo, M.; Li, C.; Zhang, R.; Yu, T.; Wang, G.; and Chen, Y. 2024.
\newblock Towards building the federatedGPT: Federated instruction tuning.
\newblock In \emph{ICASSP 2024-2024 IEEE International Conference on Acoustics, Speech and Signal Processing (ICASSP)}. IEEE.

\bibitem[{Zhuang, Chen, and Lyu(2023)}]{zhuang2023foundation}
Zhuang, W.; Chen, C.; and Lyu, L. 2023.
\newblock When foundation model meets federated learning: Motivations, challenges, and future directions.
\newblock \emph{arXiv preprint arXiv:2306.15546}.

\end{thebibliography}

\end{document}